\documentclass[runningheads]{llncs}
\usepackage{amsmath}
\usepackage{hyperref}
\usepackage{amssymb}
\usepackage[final]{graphicx}
\usepackage{subcaption}
\usepackage{tikz}
\usetikzlibrary{matrix}
\usetikzlibrary{decorations.pathreplacing}
\usepackage{graphicx}
\usepackage{arydshln}
\usepackage{multirow}
\usepackage{cite}

\newcommand{\buddidata}{\mathtt{Med_{data}}}
\newcommand{\image}{\mathtt{Img}}
\newcommand{\tab}{\mathtt{Tab}}
\newcommand{\dexter}{\mathtt{DEXTER}}
\newcommand{\amazon}{\mathtt{Textract}}
\newcommand{\microsoft}{\mathtt{AzureFR}}

\begin{document}
\title{DEXTER: An end-to-end system to extract table contents from electronic medical health documents}
\titlerunning{DEXTER: A table content extraction system for medical documents }
\author{Nandhinee P R \and
Harinath Krishnamoorthy \and
Koushik Srivatsan \and
Anil Goyal \and
Sudarsun Santhiappan}
\institute{BUDDI.AI (A Claritrics Company), Chennai, India}

%
\maketitle              
\let\clearpage\relax
\begin{abstract}
In this paper, we propose $\dexter$, an end to end system to extract information from tables present in medical health documents, such as electronic health records (EHR) and explanation of benefits (EOB). $\dexter$ consists of four sub-system stages:- \textit{i)} table detection \textit{ii)} table type classification \textit{iii)} cell detection; and \textit{iv)} cell content extraction. 
We propose a two-stage transfer learning-based approach using CDeC-Net architecture along with Non-Maximal suppression for table detection.
We design a conventional computer vision-based approach for table type classification and cell detection using parameterized kernels based on image size for detecting rows and columns. 
Finally, we extract the text from the detected cells using pre-existing OCR engine Tessaract. 
To evaluate our system, we manually annotated a sample of the real-world medical dataset (referred to as $\buddidata$) consisting of wide variations of documents (in terms of appearance) covering different table structures, such as bordered, partially bordered, borderless, or coloured tables.
We experimentally show that $\dexter$ outperforms the commercially available Amazon Textract and Microsoft Azure Form Recognizer systems on the annotated real-world medical dataset.  

\keywords{Electronic Health Records\and EHR \and Explanation of Benefits \and EOB \and Revenue Cycle Management \and Table Detection \and Cell Detection \and Table Type Classification \and Content Extraction \and RCM}
\end{abstract}
\section{Introduction}
\label{sec: intro}
With the adoption of electronic data in the healthcare space, there is an increase in demand to find the best ways to extract relevant information from the documents to help various stakeholders, such as doctors, patients, hospitals, and insurance companies. 
Apart from text, the tables present in electronic health records (EHRs) contain useful information like clinical analysis and laboratory results which are useful for research, medical investigations, clinical support system, and quality improvement.
In this paper, we propose an end-to-end system named  as \textbf{$\dexter$} (\textbf{D}ocument \textbf{Ext}ractor) which automatically extracts the data from tables  present in medical documents. \\
\newline
\textbf{Related Work.} 
In the literature, many studies have been conducted to address the challenges in content extraction from tables present in documents.
Most of the proposed approaches can be categorized into two groups depending on the type of algorithms used to deal with the problem.
Computer-vision based approaches \cite{kasar2013learning,ghanmi2015separator,shi2013model,adamo2015automatic} mainly focus on detecting lines or white patches between rows or columns in tables.
Kasar \textit{ et al.} \cite{kasar2013learning} trained an SVM classifier using lines present in the scanned documents to classify table regions.
However, the proposed method is only applicable to bordered or partially-bordered tables.
Ghanmi \textit{ et al.} \cite{ghanmi2015separator} designed an algorithm to detect lines from bordered tables.
However, for the documents containing borderless tables, the proposed approach relies on finding out the inherent syntax of the table's content. 
Given the wide variations in the appearance of medical documents, it is impossible to find the content based syntax for all kinds of documents. 
Shi \textit{et al.} \cite{shi2013model} identified table candidates from a fixed list of table models represented by a matrix of horizontal and vertical lines. 
For medical documents, Adamo \textit{et al.} \cite{adamo2015automatic} proposed a conventional image processing based method to detect tables from laboratory reports which had fixed document structure.  
However, in this work, we are interested in developing a generic system applicable to a wide range of medical documents and table structures.


With the surge in deep learning, various CNN based architectures~\cite{gilani2017table,hao2016table,li2019tablebank,paliwal2019tablenet,schreiber2017deepdesrt,prasad2020cascadetabnet,xue2018table} have been proposed for table detection and cell extraction.
Hao \textit{et al.} \cite{hao2016table} used a combination of heuristic rules with the CNN model to determine table-like structures and classify them into table or non-table regions. 
Schreiber \textit{et al.} \cite{schreiber2017deepdesrt} proposed the deep transfer learning-based algorithm DeepDeSRT for table detection and table structure recognition.
Concretely, DeepDeSRT fine-tunes a pre-trained Faster RCNN model\cite{ren2015faster} and  FCN segmentation model \cite{long2015fully}  for table detection and table structure recognition, respectively.
Li \textit{et al.} \cite{li2019tablebank}  used Faster R-CNN with ResNeXt\cite{xie2016aggregated} as the backbone architecture for table detection.
Recently, Prasad \textit{et al.}\cite{prasad2020cascadetabnet} designed an end to end approach, CascadeTabNet, using a single CNN model (utilizing iterative transfer learning) for both table detection and table segmentation. 
 For medical laboratory reports, Xue \textit{ et al.} \cite{xue2018table} proposed end-to-end system for table detection and information extraction from laboratory reports using a CNN-based model.
However, the proposed approach is only applicable to a fixed variety of documents, i.e. laboratory reports with a fixed structure and  with fixed number of tables. 
Whereas in medical domain, it is common to have a wide variety of documents with different tabular layouts and structures (bordered, partially bordered, borderless or coloured tables). 
Therefore, our objective is to design a generic end to end system that can extract content from the tables present in medical documents. 
Amazon's Textract\footnote{\url{https://aws.amazon.com/textract/}} and Microsoft's Form Recognizer\footnote{\url{https://azure.microsoft.com/en-in/services/cognitive-services/form-recognizer/}} are generic end to end commercial pipelines for information extraction from tables in a given document. 
\begin{figure}
    \centering
    \includegraphics[scale=0.45]{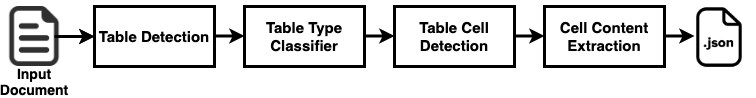}
    \caption{The pipeline of the proposed $\dexter$ System}
    \label{fig:dexter_system}
\end{figure}
In our work, we empirically compare our system with Amazon's Textract and Microsoft's  Form Recognizer.\\
\newline
\textbf{Contribution.}  We propose an end to end system, \textbf{$\dexter$} (\textbf{D}ocument \textbf{Ext}ractor), which automatically extract the content from tables for any given input medical document, as shown in Figure \ref{fig:dexter_system}.
For any given input document, $\dexter$ System uses \textit{i)} two-stage transfer learning using CDecNet \cite{agarwal2020cdec} architecture along with Non Maximal Suppression (NMS)\cite{neubeck2006efficient} for table detection; \textit{ii)} a conventional computer vision based approaches for table type classifier and cell detection in table; and \textit{iii)} finally from detected cells, we can use any OCR engine to extract information from the cells and return the output in JSON format. 
To evaluate the proposed system, due to the unavailability of medical data, we have curated a real-world medical dataset (referred as $\buddidata$\footnote{We will release the dataset to the research community.}) consisting of $1167$ real world medical documents with wide variations in appearance and different tabular structures (bordered, partially bordered, borderless and coloured tables).

We experimentally demonstrate that, compared to the existing state-of-art approaches (FR-RNX-101\cite{xie2016aggregated} and CascadeTabNet\cite{prasad2020cascadetabnet}), CDeC-Net performs significantly better on real-world medical dataset $\buddidata$  and has better generalization abilities. 
Therefore, for table detection in medical documents, we propose to use CDeC-Net architecture, which consists of cascade Mask R-CNN\cite{cai2019cascade} with ResNeXt-101 dual backbone having deformable convolution, to detect tables present in the documents.
After initializing the network with MS COCO weights, we fine-tune it on TableBank dataset \cite{li2019tablebank} in stage 1, followed by fine-tuning on $\buddidata$ dataset in stage 2. 
Finally, to select a single table prediction from multiple overlapping predictions, we used Non Maximal Suppression (NMS)\cite{neubeck2006efficient} approach. 
We experimentally show that the two-stage deep transfer learning using CDeC-Net combined with NMS approach is an effective strategy to deal with table detection problem in medical datasets. 
The proposed method performs significantly better than Amazon's Textract and Microsoft's Form Recognizer tools.  
Instead of choosing complex deep learning based approaches, following Occam's Razor principle, we designed simple computer vision based approaches for table type classifier and cell detection. 
For the table type classifier, we propose to use parameterized horizontal and vertical kernels (based on image size) for line detection. Based on the detected lines and colour of the image, we classify tables into four categories: bordered, partially-bordered, borderless and coloured tables. 
For cell detection in borderless and partially bordered tables, we propose to find vertical and horizontal column separators (or in other words, white patches) using parameterized kernels (based on the size of the image).
Compared to Amazon's Textract and Microsoft's Form Recognizer systems, our proposed method performs significantly better on the medical dataset $\buddidata$.
Moreover, we performed root cause analysis for the lower performance of existing systems on the medical dataset.
We found out experimentally that the existing systems are not robust against the borderless tables, which is the common use-case in medical documents. \\
\newline
\noindent \textbf{Paper Organization.} In the next section, we present the proposed $\dexter$ System. Before concluding in Section 4, we present the obtained experimental
results using our approach in Section \ref{sec:experiments}.  
\section{The Proposed $\dexter$ System}
\label{sec:dexter}

\begin{figure}
\begin{subfigure}{.3\textwidth}
  \centering
  \includegraphics[scale=0.15]{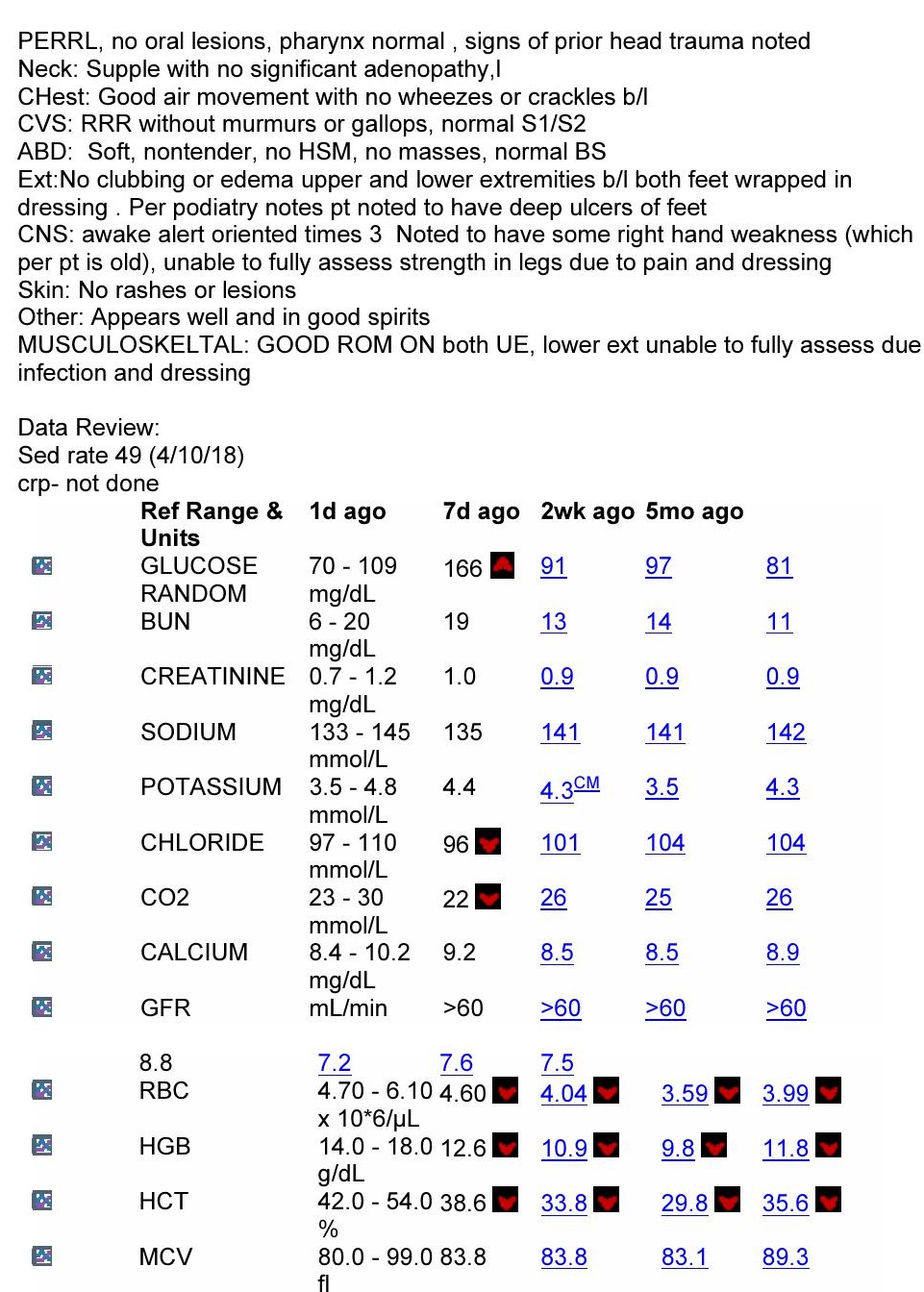}
  \label{fig:variation_borderless}
\end{subfigure}
\begin{subfigure}{.32\textwidth}
  \centering
  \includegraphics[width=0.9\linewidth]{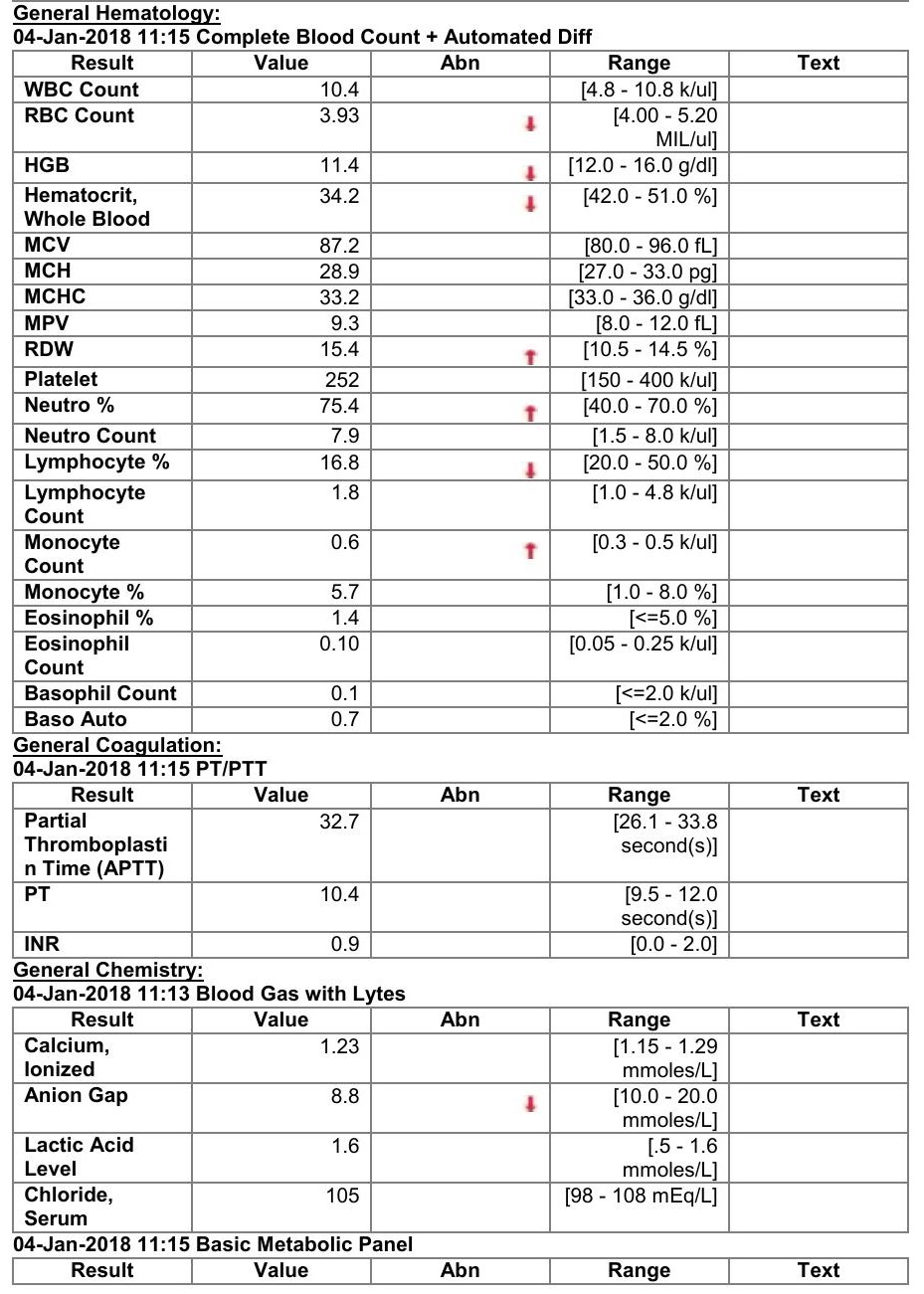}
  \label{fig:variation_bordered}
\end{subfigure}
\begin{subfigure}{.32\textwidth}
  \centering
  \includegraphics[width=0.9\linewidth]{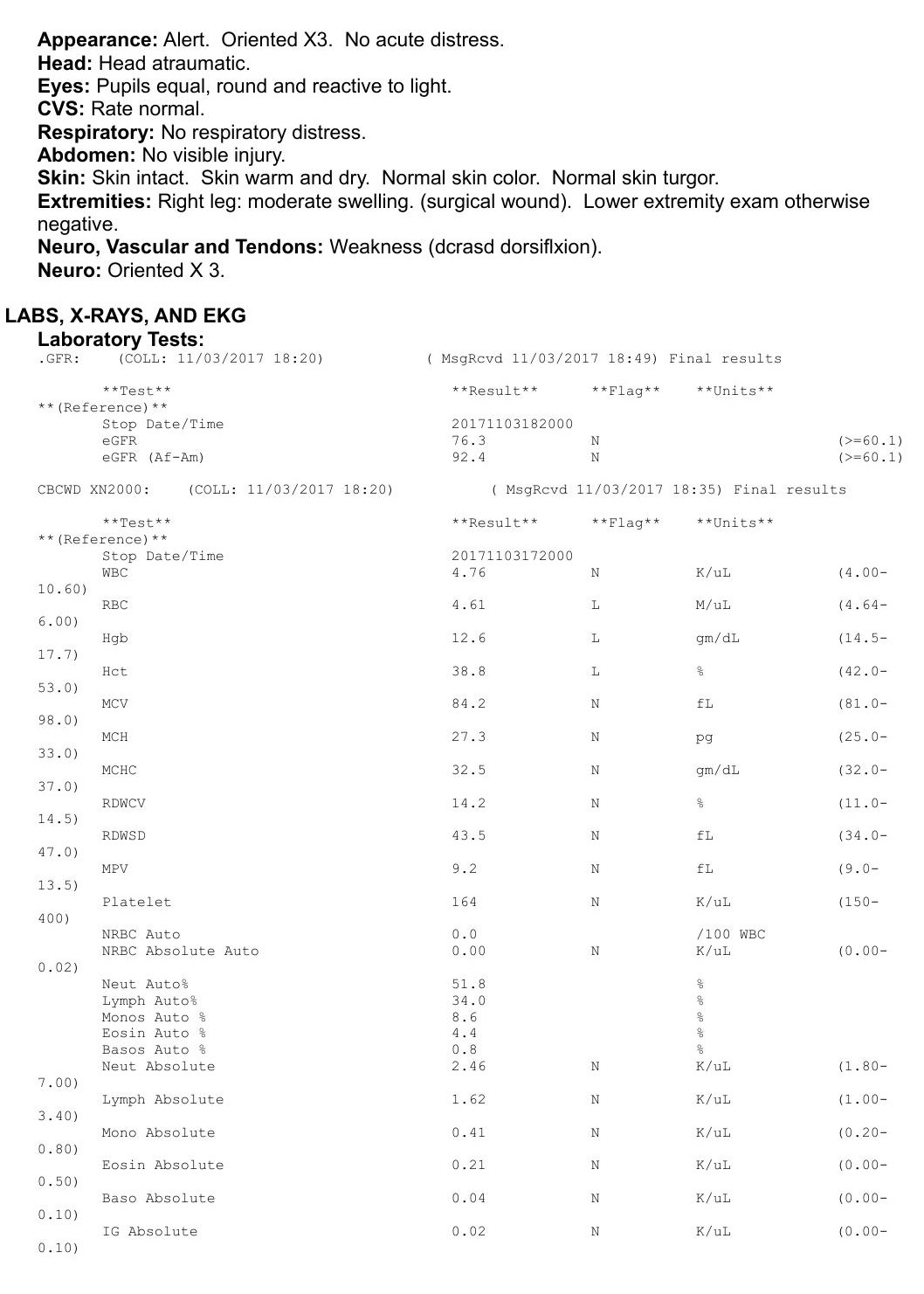}
  \label{fig:variation_spanningrow}
\end{subfigure}
\caption{Different document variations encountered in the medical domain. \textbf{Left:} investigation report. \textbf{Middle:} blood test report. \textbf{Right:} lab report.}
\label{fig:all_variations}
\end{figure}

Different types of documents are encountered in the medical domain, and a few of these include diagnostic reports, discharge summary,  prescription, case sheets, investigation reports and blood test reports. 
These documents have varied layouts and table structures, as shown in Figure \ref{fig:all_variations}. To the best of our knowledge, existing systems for extracting content from tables do not generalize well across wide variations of documents in the medical domain. 

In our work, we design a generic end to end system, referred as $\dexter$, which extracts content from varied table structures present in electronic health documents. 
For any given input medical document image $\image$, the $\dexter$ system's objective is to return the content present in the document's tables in JSON format (as shown in Figure \ref{fig:dexter_system}).
The table detection module returns co-ordinates for the detected table, $\tab$, along with the prediction confidence.
For the detected table, the table type classifier returns the table type   with output space defined as $\mathcal{Y} = \{ \text{bordered, partially bordered, borderless or coloured table} \} $. 
Based on the predictions of the table type classifier, the cell detection module returns the list of co-ordinates for detected cells within the table.
Finally, we use the existing OCR engine Tesseract\cite{smith2007overview} to extract the content of detected cells in tables.  In the next sections, we present each of these sub-modules in details.




\subsection{Table Detection}

In this section, we present a  two-stage transfer learning approach using CDeC-Net \cite{agarwal2020cdec} combined with Non Maximal Suppression (NMS)\cite{neubeck2006efficient} for detecting tables (bordered, partially bordered or borderless) in wide variations of the documents for the medical domain.

\begin{figure}[!t]
\begin{subfigure}{.5\textwidth}
  \centering
  \includegraphics[scale=0.14]{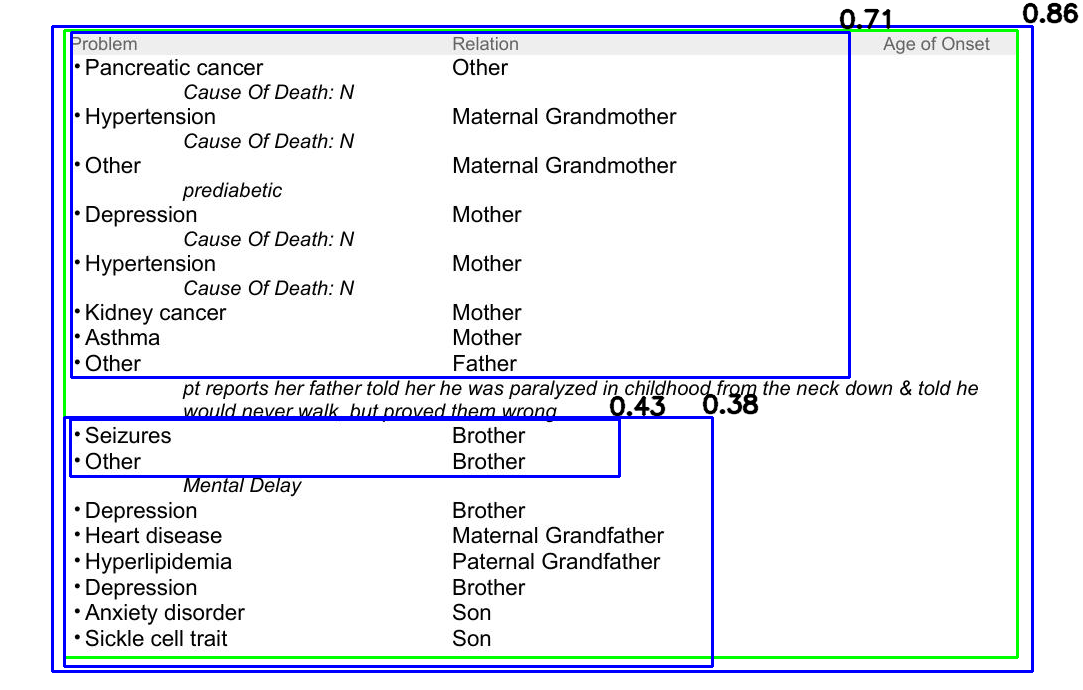}
  \label{fig:before_nms}
\end{subfigure}
\begin{subfigure}{.5\textwidth}
  \centering
  \includegraphics[scale=0.14]{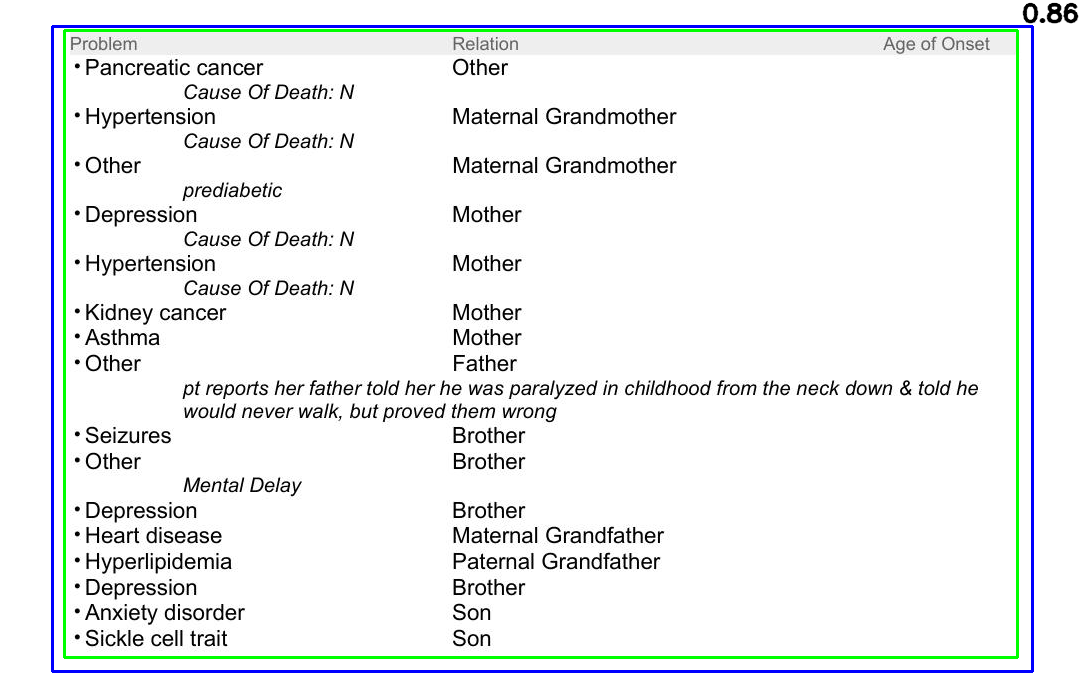}
  \label{fig:after_nms}
\end{subfigure}
\caption{Applying NMS to predictions made by CDeCNet. Green and Blue coloured rectangles correspond to ground truth and predicted bounding boxes respectively. Confidence of the prediction is included at the top-right corner of the bounding box. \textbf{Left:} Image before applying NMS depicting overlapping predictions. \textbf{Right:} Image after applying NMS with single prediction selected out of the overlapping ones.}
\label{fig:applying_nms}
\end{figure}

While dealing with table detection in medical documents, we face three major challenges. 
Firstly, we have wide variations of document structures (diagnostic reports, discharge summary, etc. ) and tables types (as shown in the Figure \ref{fig:all_variations}). 
Therefore, we need to have an approach that can be generalized to different variations and structures of documents. 
Secondly, it is common to have tables at different scales in medical documents.
Lastly, it is important to select a single table prediction from a set of falsely predicted sub-tables from an image (as shown in the left image of Figure \ref{fig:applying_nms}). 

To handle the first two challenges, we propose to use two stage transfer learning using CDeC-Net architecture.  
Concretely, we initialize the network with MS COCO weights followed by fine-tuning the network on TableBank and curated $\buddidata$ respectively.
CDeC-Net architecture uses a dual backbone, one the assistant and the other, the lead, with composite connections between the two, forming a robust backbone. 
Here, the high-level features learnt from the assistant backbone are fed as an input to the lead backbone. 
This powerful backbone helps us to handle the wide variations of documents and increase the performance of the object detector. 
Moreover, we also show experimentally, in Section \ref{sec:experiments}, that the CDeC-Net architecture has better generalization ability compared to other approaches (FR-RNX-101\cite{xie2016aggregated} and CascadeTabNet\cite{prasad2020cascadetabnet}). 
Also, to have scale-invariant table detection, CDeC-Net uses deformable CNNs that ensures that the receptive field is adaptive according to the scale of the object, thus ensuring that tables at all scales are captured correctly.

As shown in Figure \ref{fig:applying_nms} for CDeC-Net predictions, it is often possible to have multiple false sub-tables within a single tabular structure. 
Therefore, we propose to apply Non Maximal Suppression(NMS) technique. 
Concretely, NMS first selects the prediction having the highest confidence. It then computes the intersection over Union (IoU) of the selected prediction with every other prediction and discards those having IoU greater than a given threshold.
Since tables don't overlap each other, we set the threshold to $0.01$.
This is done recursively until all predictions in the image are covered. 



\subsection{Table Type Classifier}
\label{subsec:table_type_classifier}
Table type classifier is an integral sub-module of $\dexter$ pipeline because the processing steps for downstream tasks (cell detection and cell content extraction) depend on the type of table (bordered, borderless, partial bordered, coloured). 


Following Occam's Razor Principle, instead of choosing deep learning based approach, we designed a computer vision based method for table type classification. 
Concretely, we designed parameterized horizontal and vertical kernels to detect lines in a given table image. 
We classify the tables into three different categories based on the detected lines: bordered, borderless and partially bordered.



\subsubsection{Parameterized Horizontal and Vertical Kernels.}
\begin{figure}[t]
\begin{subfigure}{.5\textwidth}
  \centering
  \includegraphics[width=0.9\linewidth]{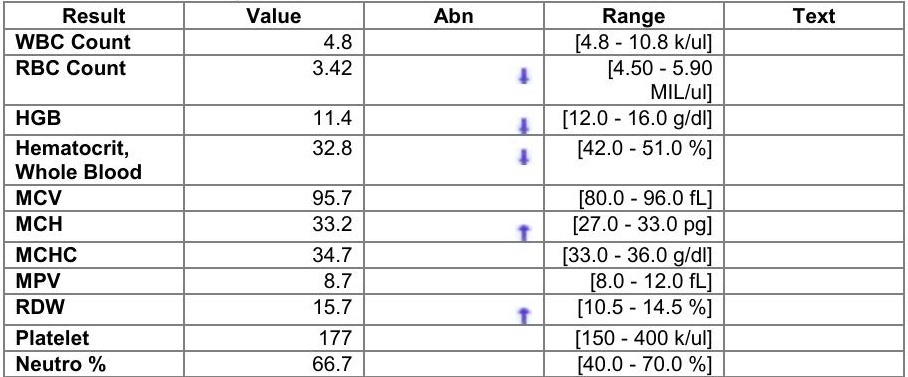}
    \caption{Original image $\tab$}
  \label{fig:table_orginal}
\end{subfigure}
\begin{subfigure}{.5\textwidth}
  \centering
  \includegraphics[width=0.9\linewidth]{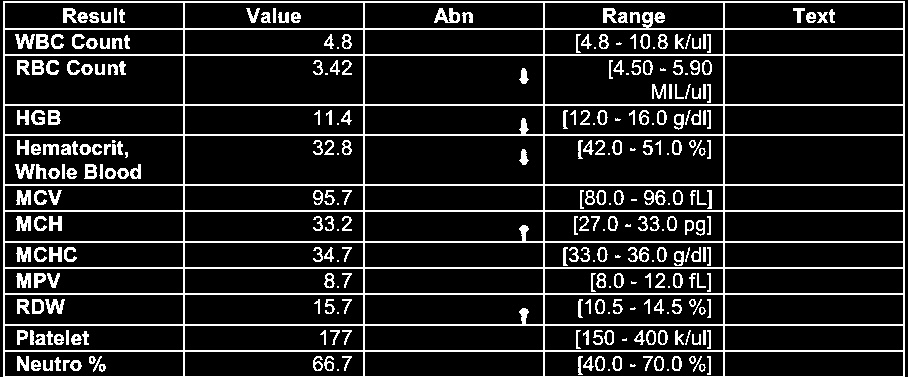}
    \caption{Thresholded and inverted image $\tab'$}
  \label{fig:table_inverted}
\end{subfigure}
\begin{subfigure}{.5\textwidth}
  \centering
  \includegraphics[width=0.9\linewidth]{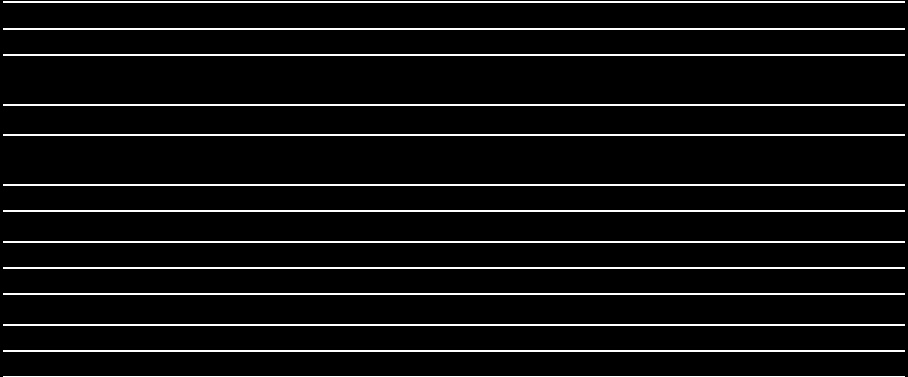}
  \caption{Image with horizontal lines $\tab_{\mathtt{hr}}$}
  \label{fig:table_horizontal_lines}
\end{subfigure}
\begin{subfigure}{.5\textwidth}
  \centering
  \includegraphics[width=0.9\linewidth]{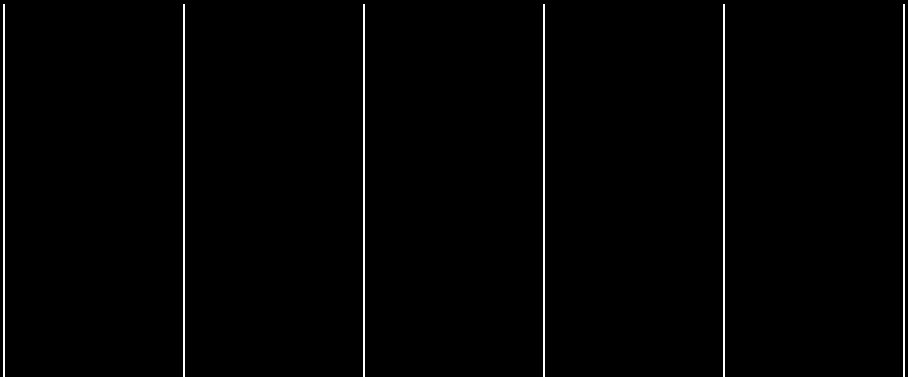}
  \caption{Image with vertical lines $\tab_{\mathtt{vr}}$}
  \label{fig:table_vertical_lines}
\end{subfigure}
\begin{subfigure}{.5\textwidth}
  \centering
  \includegraphics[width=0.9\linewidth]{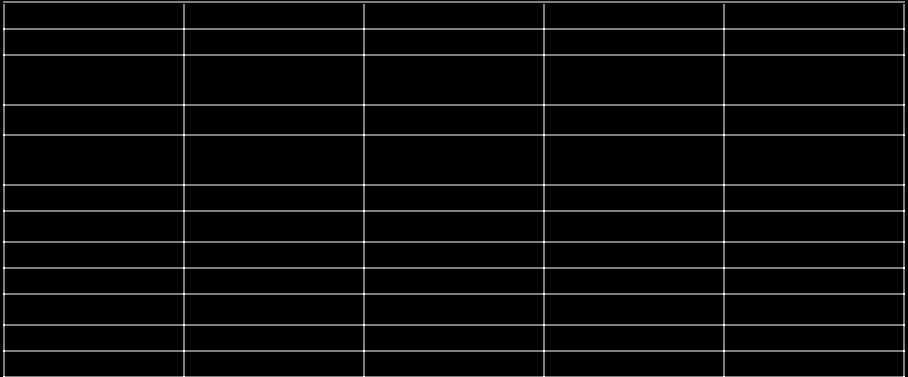}
  \caption{Image with horizontal and vertical lines $\tab_{\mathtt{lines}}$}
  \label{fig:table_both_lines}
\end{subfigure}
\begin{subfigure}{.5\textwidth}
  \centering
  \includegraphics[scale = 0.5]{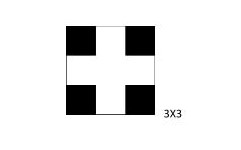}
  \caption{Template depicting Horizontal and vertical line intersection}
  \label{fig:template}
\end{subfigure}
\caption{Flow of an input image through Table type classifier module}
\label{fig:table_type_classifier}
\end{figure}

As the first step, for any given input table image  $\tab$, we apply Otsu's thresholding to the $\tab$ and invert the result to obtain $\tab'$ as shown in Figure\ref{fig:table_inverted}, which contains 1s for text region and 0s for the background region.


Instead of using fixed-sized kernels, we designed horizontal ($\mathtt{K_{hr}}$) and vertical ($\mathtt{K_{vr}}$) kernels  parameterized on size of the $\tab$  defined as follows:
 


\begin{equation}
    \mathtt{K_{hr}}  =
  \begin{bmatrix}
     1 & 1 &  1  &  \dots &  1 
  \end{bmatrix}_{\large 1 \times \mathtt{int}(\tab_w * \mathtt{K_w})}
  \label{eq:kernel_horizontal}
\end{equation}
\vspace{-1em}
\begin{equation}
    \mathtt{K_{vr}}  =
  \begin{bmatrix}
       1 & 1 &  1  &  \dots &  1 
  \end{bmatrix}_{\large \mathtt{int}(\tab_h*\mathtt{K_h}) \times  1}
  \label{eq:kernel_vertical}
\end{equation}
where, $\mathtt{K_w}$ and $\mathtt{K_h}$ are hyper-parameters kernel width and kernel height. 
Finally, we generate an image containing horizontal lines $\tab_{\mathtt{hr}}$ (Figure \ref{fig:table_horizontal_lines}) and vertical lines $\tab_{\mathtt{vr}}$ (Figure \ref{fig:table_vertical_lines}) as follows: 
\begin{equation}
    \tab_{\mathtt{hr}} = ({\tab'} \ominus \mathtt{K_{hr}}) \oplus \mathtt{K_{hr}} 
\label{eq:table_horizontal}
\end{equation}
\begin{equation}
    \tab_{\mathtt{vr}} = ({\tab'} \ominus \mathtt{K_{vr}}) \oplus \mathtt{K_{vr}} 
\label{eq:table_vertical}
\end{equation}
where ${\ominus}$ and ${\oplus}$ are erosion and dilation operations respectively.
Then, we apply hough line transform to find the number of horizontal lines ($\mathtt{Count_{hr}}$) and vertical lines ($\mathtt{Count_{vr}}$) from $ \tab_{\mathtt{hr}}$ and $ \tab_{\mathtt{vr}}$ respectively. In case we have both $\mathtt{Count_{hr}}$ and $\mathtt{Count_{vr}}$ equals to 0, then it is a borderless table.
Finally, depending on the presence of outer borders and intersection of horizontal and vertical lines we classify tables into two categories: bordered or partially bordered. 
If both outer borders and row column intersections exist, then it is bordered table, otherwise, it is a partially bordered table. 


 \vspace{-1.2em}
\subsubsection{Determining presence of outer borders.} 
We locate the top-left foreground black pixel ($\mathtt{TL_{x}}$, $\mathtt{TL_{y}}$)  and top-right foreground black pixel ($\mathtt{TR_{x}}$, $\mathtt{TR_{y}}$). If there is a line with co-ordinates (($\mathtt{TL_{x}}$, $\mathtt{TL_{y}}$), ($\mathtt{TR_{x}}$, $\mathtt{TR_{y}}$)) in $\tab_{\mathtt{hr}}$ (Equation \ref{eq:table_horizontal}), then it indicates a top border.
Similarly, we check the presence of bottom, left and right borders. 

 \vspace{-1.2em}
\subsubsection{Determining row-column intersections.} To find the row-column intersections, we define a kernel as (shown in figure \ref{fig:template}):
\begin{equation}
    \mathtt{K_{cross}} =
  \begin{bmatrix}
    0 & 1 & 0\\
     1 & 1 & 1\\
     0 & 1 & 0\\
  \end{bmatrix}_{3 \times 3}
  \label{eq:kernel_cross}
\end{equation}


We use the Hit-or-Miss Transform \cite{503921}, to find if this kernel exists in the image $\mathtt{\tab_{lines}}$ defined as: 
\begin{equation}
    \mathtt{\tab_{lines}} = \mathtt{\tab_{hr}} |  \mathtt{\tab_{vr}}
    \label{eq:table_lines}
\end{equation}
where $|$ is bitwise OR operation on images. 
A single occurrence of $ \mathtt{K_{cross}} $ signifies the presence of row-column intersection. 

$\mathtt{\tab_{lines}}$ (shown in Figure \ref{fig:table_both_lines}) is obtained by adding the images $\mathtt{\tab_{hr}}$ and $\mathtt{\tab_{vr}}$. A single occurrence of $\mathtt{K_{cross}}$ in image $\mathtt{\tab_{lines}}$ denotes the position of a row separator intersecting with a column separator.
 \vspace{-1.2em}
\subsubsection{Coloured table detection.} 
In coloured tables,  the count of foreground pixels is much higher than the count of background pixels.
Therefore, we compute the ratio of the second highest and highest intensities from histogram ($\mathtt{Hist}_{\tab}$) of grayscale table image $\tab_{\mathtt{gray}}$.
If the ratio is higher than a certain threshold $\mathtt{T}$, then it is classified as the coloured table.



\subsection{Table Cell Detection}
In this section, we present a computer vision based approach to detect cells depending upon the type of table. 
For bordered tables, we can easily find contours from  $\tab_{\mathtt{lines}}$ image (Equation \eqref{eq:table_lines}), which corresponds to cell regions in the table. 
However, for partially bordered tables, we propose to first remove the existing borders, followed by the identification of row and column separators.
Similarly, for borderless tables, we can directly identify row and column separators. 
After identifying row-column separators, we follow the same strategy for bordered tables to identify cells in partially bordered and borderless tables.



\begin{figure}[!t]
\begin{subfigure}{.49\textwidth}
  \centering
  \includegraphics[width=0.9\linewidth]{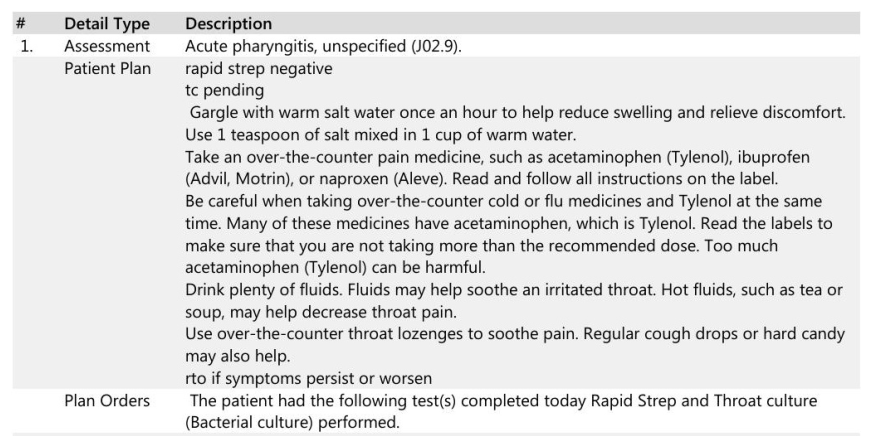}
    \caption{Original image $\tab$}
  \label{fig:table2_orginal}
\end{subfigure}
\begin{subfigure}{.49\textwidth}
  \centering
  \includegraphics[width=0.9\linewidth]{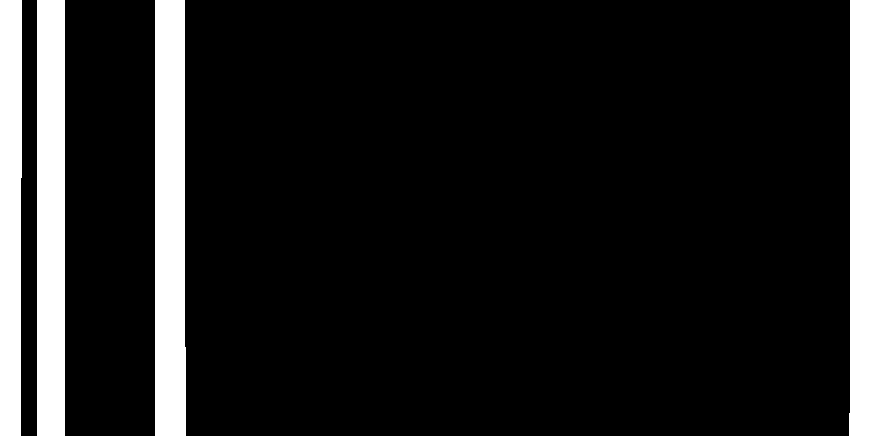}
  \caption{Image with column separators $\tab_{\mathtt{ColSeparators}}$}
  \label{fig:table2_col_sep}
\end{subfigure}
\begin{subfigure}{.49\textwidth}
  \centering
  \includegraphics[width=0.9\linewidth]{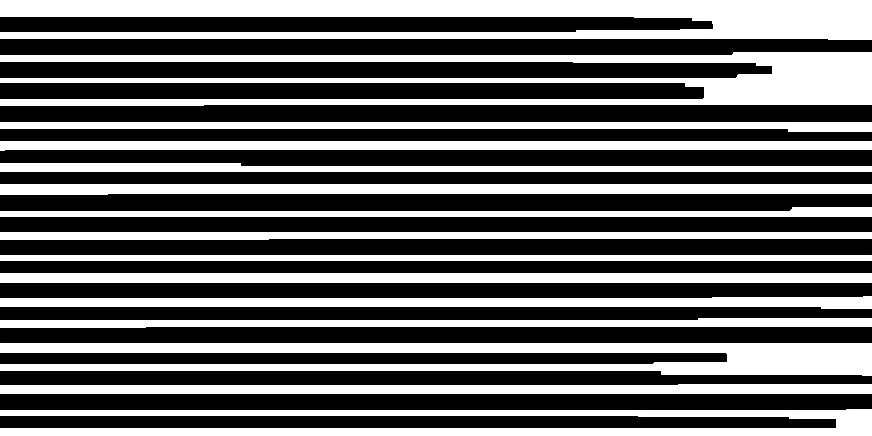}
  \caption{Image with row separators $\tab_{\mathtt{RowSeparators}}$}
  \label{fig:table2_row_sep}
\end{subfigure}
\begin{subfigure}{.49\textwidth}
  \centering
  \includegraphics[width=0.9\linewidth]{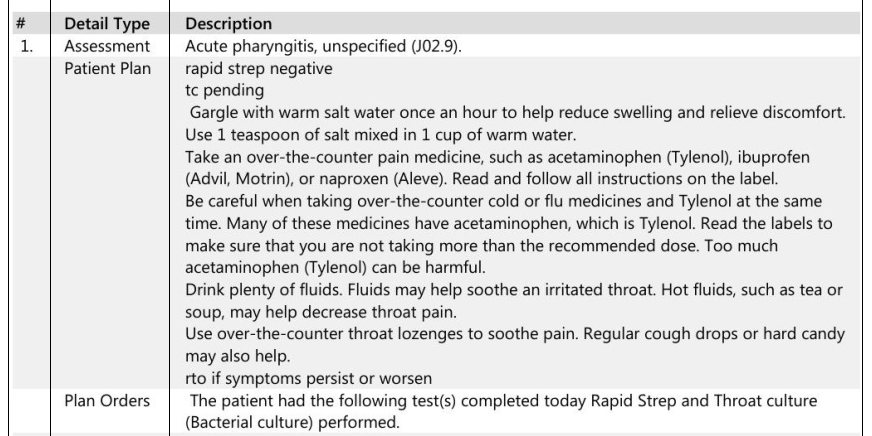}
  \caption{Image with vertical lines}
  \label{fig:table2_col_lines}
\end{subfigure}
\begin{subfigure}{.49\textwidth}
  \centering
  \includegraphics[width = 0.9\linewidth]{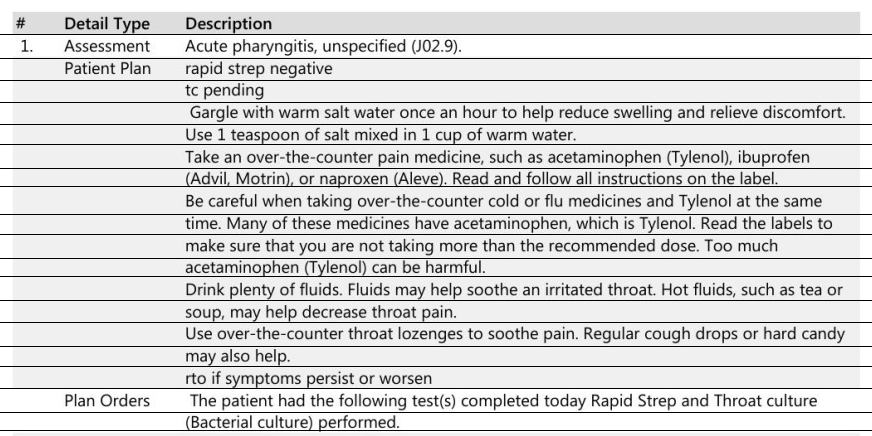}
  \caption{Image with horizontal lines}
  \label{fig:table2_row_lines}
\end{subfigure}
\begin{subfigure}{.49\textwidth}
  \centering
  \includegraphics[width = 0.9\linewidth]{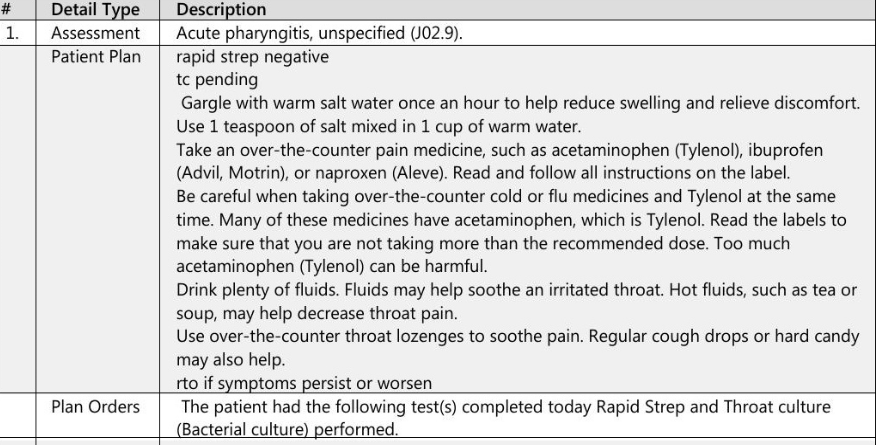}
  \caption{Final table image with horizontal and vertical lines}
  \label{fig:table2_final}
\end{subfigure}
\caption{Flow for identifying row and column separators}
\label{fig:table_cell_detection}
\end{figure}
\subsubsection{Identifying Row and Column Separators.}
\label{subsec: identify_row_col_sep}
There are two challenges while identifying row and column separators in borderless tables: \textit{i)} locating horizontal and vertical white patches in the table and \textit{ii)} handling rows which span over multiple lines of text (common in medical documents).

To handle the first challenge, we propose to use the parameterized kernels for identifying white patches in the table image $\tab$. 
As the first step, we apply Otsu's thresholding to the original table image $\tab$ which outputs an image $\mathtt{\tab_{otsu}}$  where 0s denote foreground pixels, and 1s denote background pixels.
For identifying vertical white patches, we define the vertical slider kernel  as follows:
\begin{equation}
    \mathtt{K_{vr}^{SL}} =
  \begin{bmatrix}
     1 & \dots & 1\\
     \vdots & \vdots & \vdots\\
     1 & \dots & 1\\
  \end{bmatrix}_{\mathtt{{\tab_{h}}} \times \mathtt{Sl_{w}}}
  \label{eq:slider_vertical}
\end{equation}

where $\tab_\mathtt{h}$ denotes the height of table and $\mathtt{Sl_w}$ is a  hyper-parameter computed based on the width and height of image $\tab$.
Finally, we convolve the above kernel with Otsu table image $\tab_{\mathtt{otsu}}$ as follows (see figure \ref{fig:table2_col_sep}):
\begin{equation}
    \tab_{\mathtt{ColSeparators}} = \mathtt{K_{vr}^{SL}} \circledast  \tab_{\mathtt{otsu}}
    \label{eq:table_column_separators}
\end{equation}
where $\circledast$ represents convolution operation. 
Similarly, to find the horizontal white patches in otsu's tab image $\mathtt{\tab_{otsu}}$, we define the horizontal slider kernel  as follows:
\begin{equation}
    \mathtt{K_{hr}^{SL}}   =
  \begin{bmatrix}
     1 & 1 &  1  &  \dots &  1 
  \end{bmatrix}_{\large 1\  \times\ \mathtt{\tab_{w}}}
  \label{eq:8}
\end{equation}
where $\tab_\mathtt{w}$ denotes the width of table.
Finally, we convolve the above kernel with Otsu table image $\tab_{\mathtt{otsu}}$ as follows (see figure \ref{fig:table2_row_sep}):
\begin{equation}
    \tab_{\mathtt{RowSeparators}} = \mathtt{K_{hr}^{SL}} \circledast  \tab_{\mathtt{otsu}}
    \label{eq:table_row_separators}
\end{equation}
Finally, we draw row and column separators at the middle of the white patches in vertically and horizontally convolved images $\tab_{\mathtt{ColSeparators}}$ and $\tab_{\mathtt{RowSeparators}}$ (see figures \ref{fig:table2_col_lines} and \ref{fig:table2_row_lines}). \\
\newline
To solve the second challenge (rows spanning multiple lines), it is necessary to refine the row separators in $\tab_{\mathtt{RowSeparators}}$ (Equation \ref{eq:table_row_separators}).
We propose to use the information about the number of filled cells in a given row. 
For any row in the table image, if the number of filled cells is less than the threshold  $\mathtt{cellsFilled_{thresh}}$ we remove the row separator corresponding to that row (see figure \ref{fig:table2_final}).

\subsection{Table Cell Content Extraction}

Table Cell content extraction can be facilitated using any OCR engine such as Tesseract, Abby, Microsoft Azure, etc. Note that, instead of making individual OCR calls for every cell in the table, we batch all the cells in a row and make an OCR call. This approach brings down the number of OCR calls made thereby, increasing the throughput of the system.
\section{Experiments}
\label{sec:experiments}
In this section, we present the empirical study to show the performance of $\dexter$ for three sub-modules: Table Detection, Cell Detection and Cell Content Extraction. 
Moreover, we present the root cause analysis for better performance of $\dexter$ system compared to Amazon Textract ($\amazon$) and Microsoft Azure's Form Recognizer ($\microsoft$).


\subsection{Experimental Setting}

\subsubsection{Data Preparation}
To evaluate the performance of $\dexter$ on the medical dataset, we have curated a real world medical dataset (referred as $\buddidata$) containing $1167$ images from Electronics Health Records (EHR).\footnote{The Protected Health Information (PHI) has been redacted from all the samples. }
These EHRs include wide variations of documents like investigation reports, blood test report, and discharge summary among others.\footnote{We will release the dataset to the research community.}

For any given medical image, we use the VGG Image Annotator tool \cite{dutta2019vgg}\footnote{\url{http://www.robots.ox.ac.uk/~vgg/software/via/}} to manually annotate the table and cell regions.
We export the annotated table and cell bounding box coordinates in the COCO\cite{lin2014microsoft} JSON format. 
For cell content annotations, we pass the annotated cell bounding box to an OCR for extracting the content and save the content in CSV format after manual verification. 

\vspace{-1.5em}
\subsubsection{Train Test Split}
For all our experiments, we reserve approximately $25\%$ of total images for testing and the remaining for training. 
The training and test images contain 1873 tables and 589 tables in total respectively.
To understand the performance of the systems for different table categories, we split these tables into four different table categories as shown in Table \ref{tab:train_test_cateogry_split}, with their support count. Samples containing more than one table category, are included in the splits of all relevant table categories (to maintain uniformity).
\begin{table}
    \centering
    \begin{tabular}{|c|c|c|}
        \hline
        \textbf{Table Category} & \textbf{Training dataset}  & \textbf{Testing dataset} \\
        \hline
        Bordered Tables & 249 & 163 \\ 
        \hline
        Borderless Tables & 1339 & 250 \\  
        \hline
        Partially Bordered Tables & 261 & 53 \\
        \hline
        Colour Separated Tables & 24  & 123 \\
        \hline
        Total & 1873 & 589 \\
        \hline
    \end{tabular}
    \caption{Category Wise Split of Table Count in Training and Testing Dataset}
    \label{tab:train_test_cateogry_split}
\end{table}
\subsubsection{Training Environment}
The experiments were performed on NVIDIA GeForce RTX 2080 Ti GPU with 12 GB GPU memory, Intel(R) Core(TM) i7-5930K CPU @ 3.50GHz and 32 GB of RAM.
\subsubsection{Hyper-parameter Tuning.}
For the table type classifier, we experimentally tune the hyper-parameter value $\mathtt{K_w}$ (Equation \ref{eq:kernel_horizontal}) and $\mathtt{K_h}$ (Equation  \ref{eq:kernel_vertical}) to $0.15$ and $0.1$ respectively. For colored table classification, we tune the value $\mathtt{T}$ to 0.25. Similarly, for table cell detection, we tune the value $\mathtt{Sl_w}$ based on the width and the height of the table image $\tab$. When the height of table is greater that its width,  $\mathtt{Sl_w}$ is set to 1. When the height is less that 360px, $\mathtt{Sl_w}$ is set to 4. For other cases, $\mathtt{Sl_w}$ is set to 2. 
For refining the row separators in $\tab_{\mathtt{RowSeparators}}$, we set $\mathtt{cellsFilled_{thresh}}$ to be one added with half of the number of columns in the table.

\vspace{-1em}
\subsubsection{Evaluation Metrics.}
\label{subsec: Evaluation Metrics}
As followed in the literature\cite{padilla2020survey,agarwal2020cdec,prasad2020cascadetabnet,gilani2017table,paliwal2019tablenet}, we use precision (P), recall (R), F1-score (F1), and mean average precision (mAP) to evaluate the performance of table detection and cell detection modules at multiple Intersection over Union (IoU) thresholds.
For cell content extraction, we use the edit-distance based metric at the character level, where we quantifying how dissimilar two strings are to one another by counting the minimum number of operations required to transform one string into the other. 
If the edit-distance is less than a set value (0, 2 or 3), then we consider that as a correctly  classified sample.

\subsection{Experiment Results}

\subsubsection{Preliminary Analysis.}
Several SOTA architectures have shown promising results to detect tables in a given image. In order to identify the best performing architecture for our use case, we design an experiment to determine two things: \textit{i)} performance of the architectures on the widely used TableBank \cite{li2019tablebank} dataset and \textit{ii)} generalizability of the architectures on the unseen $\buddidata$ dataset.\

We choose three architectures, namely, FR-RNX-101\cite{xie2016aggregated},  CasacadeTabNet\cite{prasad2020cascadetabnet} and CDeC-Net\cite{agarwal2020cdec}, where CasacadeTabNet and CDeC-Net are the state of the art models for table detection, and FR-RNX-101 is the current TableBank baseline model. We initialise these architectures, with their base weights provided by the authors and then fine-tune it on the train split of the TableBank dataset \cite{li2019tablebank}. Each model in Stage-1 is trained over 2 epochs. We then evaluate these three models on the test split of TableBank and $\buddidata$ datasets. \
From Table \ref{tab: stage1_results_score}, we can see that CDeC-Net outperforms both the architectures for both TableBank and $\buddidata$ test sets, thus demonstrating better generalizability. We hence choose CDeC-Net as our base architecture.
For the second stage of the two stage fine-tuning process, we take the weights of CDeC-Net trained on TableBank dataset and fine-tune on the train split of $\buddidata$ for $30$ epochs. 

\begin{table}[]
\centering
\begin{tabular}{|c|c|c|c|c|c|}
\hline
\multirow{2}{*}{\textbf{Method Name}} & \multirow{2}{*}{\textbf{Train Dataset}} & \multirow{2}{*}{\textbf{Test Dataset}} & \multicolumn{3}{c|}{\textbf{Avg. Score (0.5 - 0.95 IoU)}} \\ \cline{4-6} 
                                      &                                            &                                        & \textbf{P}     & \textbf{R}    & \textbf{F1}    \\ \hline
\multirow{2}{*}{FR-RNX-101 \cite{xie2016aggregated}}           & \multirow{2}{*}{Tablebank}                 & Tablebank                              & 0.95                  & 0.97              & 0.96          \\ \cline{3-6} 
                                      &                                            & $\buddidata$                           & 0.22                  & 0.24               & 0.23          \\ \hline
\multirow{2}{*}{CascadeTabNet\cite{prasad2020cascadetabnet}}        & \multirow{2}{*}{Tablebank}                 & Tablebank                              & 0.93                  & 0.95              & 0.94          \\ \cline{3-6} 
                                      &                                            & $\buddidata$                           & 0.33                  & 0.42              & 0.37          \\ \hline
\multirow{2}{*}{CDeC-Net\cite{agarwal2020cdec}}             & \multirow{2}{*}{Tablebank}                 & Tablebank                              & 0.96                  & 0.98              & 0.97          \\ \cline{3-6} 
                                      &                                            & $\buddidata$                           & 0.40                    & 0.47              & 0.43          \\ \hline
\end{tabular}
\caption{Preliminary Analysis score}
\label{tab: stage1_results_score}
\end{table}
\begin{figure}[htb]
\begin{subfigure}{.33\linewidth}
  \centering
  \includegraphics[scale=0.1]{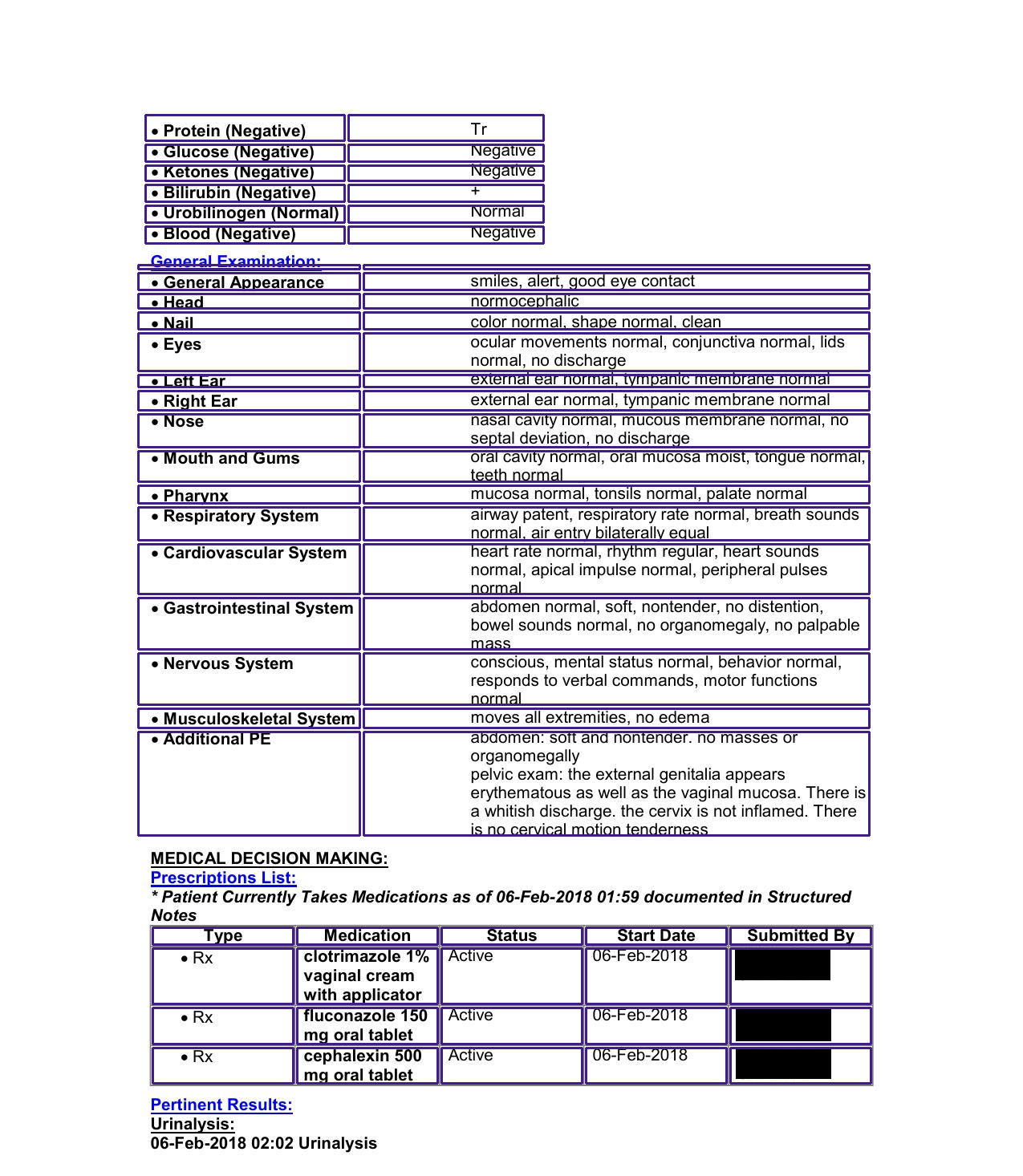}
    \caption{$\dexter$}
  \label{fig:dexter_cell_pred}
\end{subfigure}
\begin{subfigure}{.33\linewidth}
  \centering
  \includegraphics[scale=0.1]{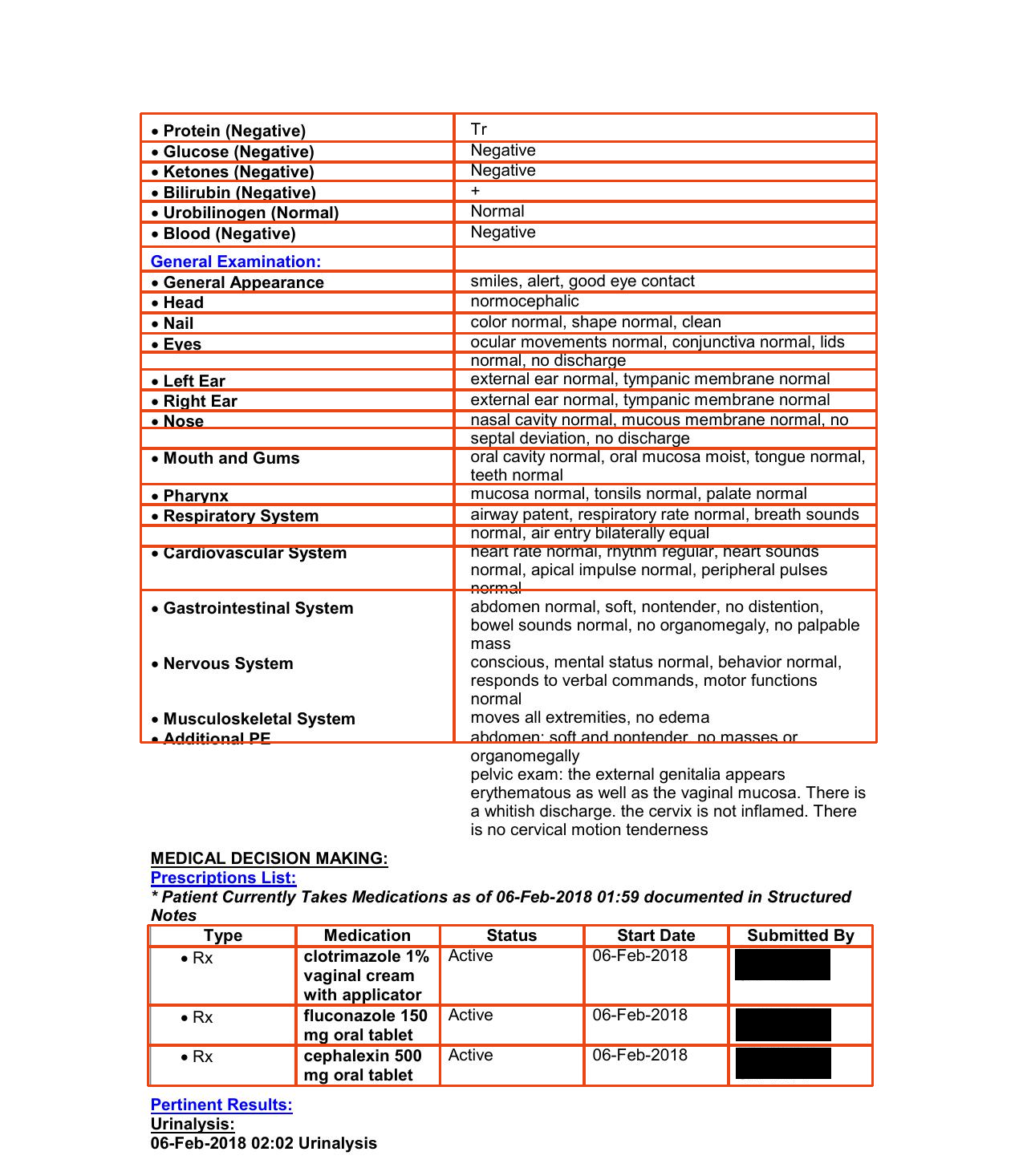}
  \caption{$\amazon$}
  \label{fig:textract_cell_pred}
\end{subfigure}%
\begin{subfigure}{.33\linewidth}
  \centering
  \includegraphics[scale=0.1]{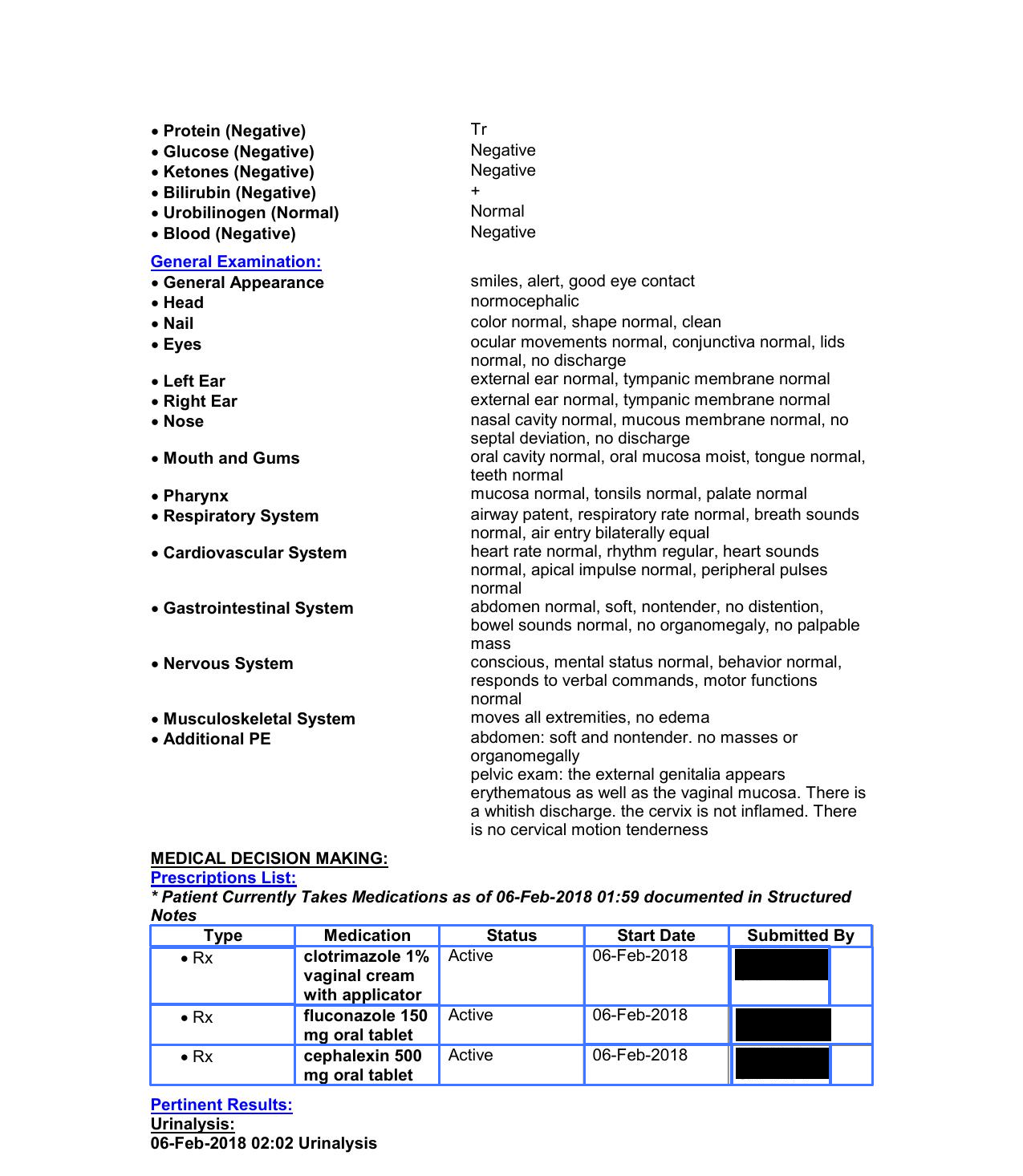}
  \label{fig:azure_cell_pred}
  \caption{$\microsoft$}
\end{subfigure}%
\caption{Table and Cell Detection predictions by $\dexter$, $\amazon$ and $\microsoft$}
\label{fig:table_and_cell_det_vendor_pred}
\end{figure}

\vspace{-2em}
\subsubsection{Table Detection.}
In Table \ref{tab:table_detection_combined_split}, we present the experimental results for the table detection module. 
From table, we can deduce that the absolute gain in performance in terms of F1-score ranges between $19\%$ to $30\%$ at different IoU thresholds.
Moreover, we can see that the proposed $\dexter$ system is a high precision and high recall model which is important in medical document processing. 
Whereas, $\amazon$ and $\microsoft$ are high precision models. 
This indicates that, $\dexter$ was able to precisely capture most of the table in the medical charts, and performs well on the pages containing multiple table (as depicted in figure  \ref{fig:table_and_cell_det_vendor_pred}), while $\amazon$ and $\microsoft$ find it difficult to predict all tables in a given page.

\begin{table}[]
\centering
\begin{tabular}{|c|c|c|c|c|c|c|c|c|c|c|c|c|}
\hline
\multirow{2}{*}{\textbf{IoU}} & \multicolumn{4}{c|}{\textbf{$\dexter$}}                           & \multicolumn{4}{c|}{\textbf{$\amazon$}}                     & \multicolumn{4}{c|}{\textbf{$\microsoft$}}                     \\ \cline{2-13} 
                              & \textbf{P} & \textbf{R} & \textbf{F1} & \textbf{mAP(\%)} & \textbf{P} & \textbf{R} & \textbf{F1} & \textbf{mAP(\%)} & \textbf{P} & \textbf{R} & \textbf{F1} & \textbf{mAP(\%)} \\ \hline
0.5                                                    & 0.88                                                            & 0.91                                                         & 0.9                                                      & 89.30\%                                                   & 0.88                                                            & 0.59                                                         & 0.71                                                     & 57.79\%                                                   & 0.97                                                            & 0.49                                                         & 0.65                                                     & 47.63\%                                                   \\ \hline
0.6                                                    & 0.88                                                            & 0.9                                                          & 0.89                                                     & 87.67\%                                                   & 0.86                                                            & 0.58                                                         & 0.69                                                     & 55.45\%                                                   & 0.95                                                            & 0.48                                                         & 0.64                                                     & 45.96\%                                                   \\ \hline
0.7                                                    & 0.85                                                            & 0.88                                                         & 0.86                                                     & 84.04\%                                                   & 0.82                                                            & 0.55                                                         & 0.66                                                     & 52.00\%                                                   & 0.91                                                            & 0.46                                                         & 0.61                                                     & 42.67\%                                                   \\ \hline
0.8                                                    & 0.8                                                             & 0.82                                                         & 0.81                                                     & 76.97\%                                                   & 0.75                                                            & 0.5                                                          & 0.6                                                      & 45.15\%                                                   & 0.82                                                            & 0.42                                                         & 0.55                                                     & 35.37\%                                                   \\ \hline
0.9                                                    & 0.69                                                            & 0.71                                                         & 0.7                                                      & 61.11\%                                                   & 0.57                                                            & 0.38                                                         & 0.46                                                     & 30.24\%                                                   & 0.59                                                            & 0.3                                                          & 0.4                                                      & 20.17\%                                                   \\ \hline

\end{tabular}
\caption{Table detection scores on the test split of $\buddidata$}
\label{tab:table_detection_combined_split}
\end{table}

\begin{table}[]
\centering
\begin{tabular}{|c|c|c|c|c|c|c|c|c|c|c|c|c|}
\hline
\multirow{2}{*}{\textbf{IoU}} & \multicolumn{4}{c|}{\textbf{$\dexter$}}                           & \multicolumn{4}{c|}{\textbf{$\amazon$}}                     & \multicolumn{4}{c|}{\textbf{$\microsoft$}}                     \\ \cline{2-13} 
                              & \textbf{P} & \textbf{R} & \textbf{F1} & \textbf{mAP(\%)} & \textbf{P} & \textbf{R} & \textbf{F1} & \textbf{mAP(\%)} & \textbf{P} & \textbf{R} & \textbf{F1} & \textbf{mAP(\%)} \\ \hline
0.5                           & 0.83                                    & 0.73                                 & 0.78                             & 61.36\%                           & 0.61                                    & 0.65                                 & 0.63                             & 43.02\%                           & 0.62                                    & 0.55                                 & 0.58                             & 36.67\%                           \\ \hline
0.6                           & 0.78                                    & 0.69                                 & 0.73                             & 54.42\%                           & 0.55                                    & 0.59                                 & 0.57                             & 35.36\%                           & 0.57                                    & 0.5                                  & 0.53                             & 31.42\%                           \\ \hline
0.7                           & 0.71                                    & 0.63                                 & 0.67                             & 46.21\%                           & 0.47                                    & 0.5                                  & 0.48                             & 26.26\%                           & 0.51                                    & 0.45                                 & 0.48                             & 25.98\%                           \\ \hline
0.8                           & 0.66                                    & 0.58                                 & 0.62                             & 39.63\%                           & 0.27                                    & 0.29                                 & 0.28                             & 9.25\%                            & 0.26                                    & 0.23                                 & 0.25                             & 7.04\%                            \\ \hline
0.9                           & 0.6                                     & 0.53                                 & 0.56                             & 33.01\%                           & 0.06                                    & 0.06                                 & 0.06                             & 0.45\%                            & 0.04                                    & 0.04                                 & 0.04                             & 0.19\%                            \\ \hline
\end{tabular}
\caption{Cell detection scores on the test split of $\buddidata$}
\label{tab:cell_detection_combined_split}
\end{table}
\vspace{-1em}
\subsubsection{Cell Detection.}
In Table \ref{tab:cell_detection_combined_split}, we present the experimental results for the cell detection module. For each of the tables predicted in the previous stage, we take the cell bounding box predictions relative to the page and compute the P, R and F1 scores at multiple IoU thresholds.
We can see that the absolute gain in performance in terms of F1-score ranges between $15\%$ to $52\%$ at different IoU thresholds.
There are two reasons for this significant better performance: \textit{i)} $\dexter$ has better table detection performance in terms of both precision and recall; and \textit{ii)} $\dexter$ is able to handle well the cells where the text spans over multiple rows as shown in Figure~\ref{fig:table_and_cell_det_vendor_pred}.

\begin{table}[]
\centering
\begin{tabular}{|c|c|c|c|c|c|c|c|c|c|}
\hline
\multirow{2}{*}{\textbf{Edit-distance}} & \multicolumn{3}{c|}{\textbf{$\dexter$}}            & \multicolumn{3}{c|}{\textbf{$\amazon$}}      & \multicolumn{3}{c|}{\textbf{$\microsoft$}}      \\ \cline{2-10} 
                                        & \textbf{P} & \textbf{R} & \textbf{F1} & \textbf{P} & \textbf{R} & \textbf{F1} & \textbf{P} & \textbf{R} & \textbf{F1} \\ \hline
0 & 0.5433 & 0.3856 & 0.4511 & 0.545  & 0.4166 & 0.4722 & 0.4388 & 0.3438 & 0.3855 \\ \hline
2 & 0.6785 & 0.4816 & 0.5633 & 0.6361 & 0.4862 & 0.5512 & 0.516  & 0.4043 & 0.4534 \\ \hline
3 & 0.7188 & 0.5101 & 0.5967 & 0.6619 & 0.5059 & 0.5735 & 0.54   & 0.4231 & 0.4744 \\ \hline
\end{tabular}
\caption{Cell content extraction scores on the test split of the $\buddidata$}
\label{tab:cell_content_extraction_combined_split_1}
\end{table}
\vspace{-1em}
\subsubsection{Cell Content Extraction.}
In Table $5$, we evaluate the performance of $\dexter$'s cell content extraction module and compare it against $\amazon$ and $\microsoft$. We take each pair of ground truth and predicted table and compute the performance using the edit-distance metric, as mentioned in section\ref{subsec: Evaluation Metrics}. We report the scores at edit-distance 0 (exact match) and at edit-distance 2 and 3 (maximum of 2 and 3 characters can be incorrect). 
We can see that  $\dexter$ has a gain of $2-12\%$ in terms of F1-score at different edit-distance settings and this behaviour reflects the performance in table detection and cell detection evaluations.

\vspace{-1.5em}
\subsubsection{Root Cause Analysis.} 
We analyzed the performance of all three systems based on the category of tables.
From Table \ref{tab:train_test_cateogry_split}, we can deduce that medical documents consist of majority of borderless tables and therefore, it is important to efficiently handle the borderless table category.
We have seen experimentally that $\dexter$ is able to perform more than $30\%$ (and $28\%$) better than $\amazon$ system (the second best baseline) for table detection (and cell detection respectively) in terms of F1-score. 
\section{Conclusion}
\label{sec:conclusion}
We presented $\dexter$, an end to end system to extract content from tables present in medical health documents. 
We evaluated our system using a manually annotated real-world medical dataset consisting of $1167$ images, which we would be releasing to the research community. This covers a wide variety of types in terms of appearance and table structures.
We experimentally showed that $\dexter$ outperforms Amazon's Textract and Microsoft Azure's Form Recognizer system on the annotated medical dataset. We scored a high absolute gain in performance in terms of F1-score, which is more than $19\%$ and $15\%$ for table detection and cell detection tasks respectively.
For the cell extraction task, we reported a gain of $2-12\%$ at different edit-distance settings. 
We also performed root cause analysis for the under-performance of the existing pipelines for medical datasets. 
Our experiments showed that the existing systems are not robust against the borderless tables, which is the common use-case in medical documents.

\bibliographystyle{splncs04}
\bibliography{biblio}

\begin{thebibliography}{10}
\providecommand{\url}[1]{\texttt{#1}}
\providecommand{\urlprefix}{URL }
\providecommand{\doi}[1]{https://doi.org/#1}

\bibitem{adamo2015automatic}
Adamo, F., Attivissimo, F., Di~Nisio, A., Spadavecchia, M.: An automatic
  document processing system for medical data extraction. Measurement
  \textbf{61},  88--99 (2015)

\bibitem{agarwal2020cdec}
Agarwal, M., Mondal, A., Jawahar, C.: Cdec-net: Composite deformable cascade
  network for table detection in document images. arXiv preprint:2008.10831
  (2020)

\bibitem{cai2019cascade}
Cai, Z., Vasconcelos, N.: Cascade r-cnn: high quality object detection and
  instance segmentation. IEEE transactions on pattern analysis and machine
  intelligence  (2019)

\bibitem{dutta2019vgg}
Dutta, A., Zisserman, A.: The {VIA} annotation software for images, audio and
  video. In: Proceedings of the 27th ACM International Conference on
  Multimedia. MM '19, ACM, New York, NY, USA (2019).
  \doi{10.1145/3343031.3350535}, \url{https://doi.org/10.1145/3343031.3350535}

\bibitem{ghanmi2015separator}
Ghanmi, N., Belaid, A.: Separator and content based approach for table
  extraction in handwritten chemistry documents. In: 13th ICDAR Conference. pp.
  296--300. IEEE (2015)

\bibitem{gilani2017table}
Gilani, A., Qasim, S.R., Malik, I., Shafait, F.: Table detection using deep
  learning. In: 14th ICDAR Conference. vol.~1, pp. 771--776 (2017)

\bibitem{hao2016table}
Hao, L., Gao, L., Yi, X., Tang, Z.: A table detection method for pdf documents
  based on convolutional neural networks. In: 12th IAPR Workshop on Document
  Analysis Systems (DAS). pp. 287--292. IEEE (2016)

\bibitem{kasar2013learning}
Kasar, T., Barlas, P., Adam, S., Chatelain, C., Paquet, T.: Learning to detect
  tables in scanned document images using line information. In: 12th ICDAR
  Conference. pp. 1185--1189 (2013)

\bibitem{503921}
{Khosravi}, M., {Schafer}, R.W.: Template matching based on a grayscale
  hit-or-miss transform. IEEE Transactions on Image Processing  \textbf{5}(6),
  1060--1066 (1996). \doi{10.1109/83.503921}

\bibitem{li2019tablebank}
Li, M., Cui, L., Huang, S., Wei, F., Zhou, M., Li, Z.: Tablebank: A benchmark
  dataset for table detection and recognition (2019)

\bibitem{lin2014microsoft}
Lin, T.Y., Maire, M., Belongie, S., Hays, J., Perona, P., Ramanan, D.,
  Doll{\'a}r, P., Zitnick, C.L.: Microsoft coco: Common objects in context. In:
  ECCV. pp. 740--755. Springer (2014)

\bibitem{long2015fully}
Long, J., Shelhamer, E., Darrell, T.: Fully convolutional networks for semantic
  segmentation. In: Proceedings of the IEEE conference on computer vision and
  pattern recognition. pp. 3431--3440 (2015)

\bibitem{neubeck2006efficient}
Neubeck, A., Van~Gool, L.: Efficient non-maximum suppression. In: 18th ICPR
  Conference. vol.~3, pp. 850--855 (2006)

\bibitem{padilla2020survey}
Padilla, R., Netto, S.L., da~Silva, E.A.: A survey on performance metrics for
  object-detection algorithms. In: 2020 International Conference on Systems,
  Signals and Image Processing (IWSSIP). pp. 237--242. IEEE (2020)

\bibitem{paliwal2019tablenet}
Paliwal, S.S., Vishwanath, D., Rahul, R., Sharma, M., Vig, L.: Tablenet: Deep
  learning model for end-to-end table detection and tabular data extraction
  from scanned document images. In: ICDAR Conference. pp. 128--133 (2019)

\bibitem{prasad2020cascadetabnet}
Prasad, D., Gadpal, A., Kapadni, K., Visave, M., Sultanpure, K.: Cascadetabnet:
  An approach for end to end table detection and structure recognition from
  image-based documents. In: Proceedings of IEEE CVPR Workshops. pp. 572--573
  (2020)

\bibitem{ren2015faster}
Ren, S., He, K., Girshick, R., Sun, J.: Faster r-cnn: Towards real-time object
  detection with region proposal networks. arXiv :1506.01497  (2015)

\bibitem{schreiber2017deepdesrt}
Schreiber, S., Agne, S., Wolf, I., Dengel, A., Ahmed, S.: Deepdesrt: Deep
  learning for detection and structure recognition of tables in document
  images. In: 14th ICDAR Conference. vol.~1, pp. 1162--1167. IEEE (2017)

\bibitem{shi2013model}
Shi, Z., Setlur, S., Govindaraju, V.: A model based framework for table
  processing in degraded document images. In: 12th ICDAR Conference. pp.
  963--967. IEEE (2013)

\bibitem{smith2007overview}
Smith, R.: An overview of the tesseract ocr engine. In: 9th ICDAR. vol.~2, pp.
  629--633 (2007)

\bibitem{xie2016aggregated}
Xie, S., Girshick, R.B., Doll{\'a}r, P., Tu, Z., He, K.: Aggregated residual
  transformations for deep neural networks. corr abs/1611.05431 (2016). arXiv
  preprint arXiv:1611.05431  (2016)

\bibitem{xue2018table}
Xue, W., Li, Q., Zhang, Z., Zhao, Y., Wang, H.: Table analysis and information
  extraction for medical laboratory reports. In: 2018 IEEE 16th Intl Conf on
  Dependable, Autonomic and Secure Computing, 16th Intl Conf on Pervasive
  Intelligence and Computing, 4th Intl Conf on Big Data Intelligence and
  Computing and Cyber Science and Technology Congress
  (DASC/PiCom/DataCom/CyberSciTech). pp. 193--199. IEEE (2018)

\end{thebibliography}

\end{document}